\documentclass{article}
\usepackage{spconf,amsmath,graphicx,caption}
\usepackage{subcaption}
\usepackage{multirow}

\title{SAT: Self-adaptive training for fashion compatibility prediction}
\name{Ling Xiao and Toshihiko Yamasaki}
\address{Department of Information and Communication Engineering, \\
Graduate School of Information Science and Technology, \\
The University of Tokyo, Tokyo, Japan }
\begin{document}
%
\maketitle
\begin{abstract}
This paper presents a self-adaptive training (SAT) model for fashion compatibility prediction. It focuses on the learning of some hard items, such as those that share similar color, texture, and pattern features but are considered incompatible due to the aesthetics or temporal shifts. Specifically, we first design a method to define hard outfits and a difficulty score (DS) is defined and assigned to each outfit based on the difficulty in recommending an item for it. Then, we propose a self-adaptive triplet loss (SATL), where the DS of the outfit is considered. Finally, we propose a very simple conditional similarity network combining the proposed SATL to achieve the learning of hard items in the fashion compatibility prediction. Experiments on the publicly available Polyvore Outfits and Polyvore Outfits-D datasets demonstrate our SAT's effectiveness in fashion compatibility prediction. Besides, our SATL can be easily extended to other conditional similarity networks to improve their performance.

\end{abstract}
\begin{keywords}
Fashion compatibility prediction, fashion recommendation, adaptive loss, conditional similarity networks.
\end{keywords}
\section{Introduction}
\label{sec:intro}

\begin{figure}[t!]
\begin{center}
\includegraphics[width=0.85\linewidth]{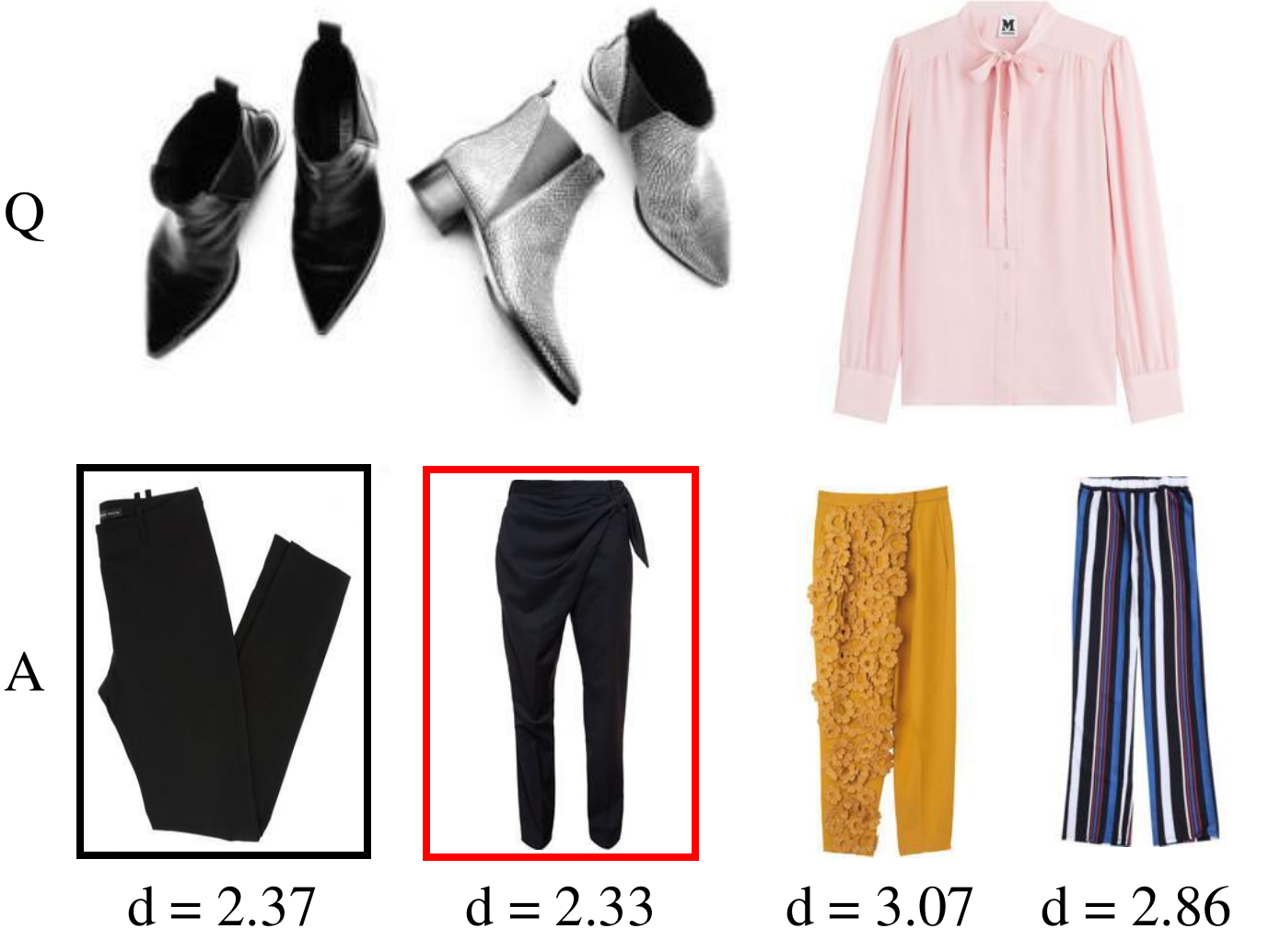}
\end{center}
   \caption{A hard example in fashion compatibility prediction. Row $Q$ is the given incomplete outfit, while row $A$ is the answer. $d$ denotes the distance of the answer item with the given incomplete outfit. The model selects the item with the lowest $d$ as the predicted answer. The item highlighted in the black box is the ground truth while the item highlighted in the red box is the false prediction result.}
\label{fig:Problem}
\end{figure}

Deep learning has achieved great success in fashion compatibility prediction~\cite{Han_ACMMM_17,Vasileva_ECCV_18,Tan_ICCV_19,Kim_ICCV_21,Lin_CVPR_20,Polania_ICIP_19,Zhan_ICIP_21}. Existing methods mainly use graph neural networks~\cite{Liu_TMM_20,Su_ACMMM_21}, conditional similarity networks~\cite{Vasileva_ECCV_18,Tan_ICCV_19,Lin_CVPR_20,Zheng_ACMMM_21,Polania_ICIP_19,Gao_Multi_Systems_19}, LSTM~\cite{Han_ACMMM_17,Nakamura_arxiv_18,Pang_CVPR_21}, and self-supervised or semi-supervised methods~\cite{Kim_ICCV_21,Revanur_RecSys_21} to directly build multiple embeddings from the image, which are used to measure the compatibility among fashion items. By using attention techniques~\cite{Lin_CVPR_20}, text
information~\cite{Vasileva_ECCV_18}, and some pretext tasks~\cite{Kim_ICCV_21,Revanur_RecSys_21}, they could achieve good results on the easy outfit. Among these methods, conditional similarity networks are mostly investigated~\cite{Vasileva_ECCV_18,Tan_ICCV_19,Lin_CVPR_20,Zheng_ACMMM_21,Polania_ICIP_19,Gao_Multi_Systems_19}. By mapping item general features to multiple embedding spaces, they can analyze complex compatibility among fashion items from different categories. However, existing methods do not take hard items into consideration, such as fashion items that share similar color, texture, and pattern features but are considered incompatible due to the aesthetics or temporal shifts (see Fig.~\ref{fig:Problem}).

To solve this problem, we present a self-adaptive training (SAT) model to let the Convolution Neural Network (CNN) learn more from the hard samples. The key insight of our approach is that the embedding distance between positive pairs ($D_p$) and negative pairs ($D_n$) should meet $D_n-D_p > \mu$. For some hard items that $D_n-D_p < \mu$, giving a higher weight to the hard items will make the CNN learn more from the hard items. Specifically, we first design a method to define hard outfits in the fashion compatibility prediction task and a difficulty score (DS) is defined and assigned to each outfit. Then, we introduce a self-adaptive triplet loss (SATL) in the embedding space, which helps to assign weights to the triplet input based on the DS. Finally, combining our SATL, we develop a very simple SAT model for fashion compatibility prediction. Our SAT only uses image features and requires no attention techniques, but can yield state-of-the-art performance. Further, our SATL can be extended to any other conditional similarity networks for improving their performance. Our SATL could also be a good inspiration for other work that focuses on learning a more discriminative embedding space.

Our contributions are summarized below:
\begin{itemize}
  \item [1)]
 To the best of our knowledge, this is the first work to propose a model to let the CNN learn more from the hard items in fashion compatibility prediction.
  \item [2)]
  We propose a SATL to learn a more discriminative embedding space.
  \item [3)]
  Our SATL can be easily extended to other condition similarity networks to improve their performance.
\end{itemize}

Our source code will be made publicly available upon acceptance.

\section{Method}
\label{sec:method}
\subsection{Overview}
\begin{figure*}[t]
\begin{center}
\includegraphics[width=0.75\linewidth]{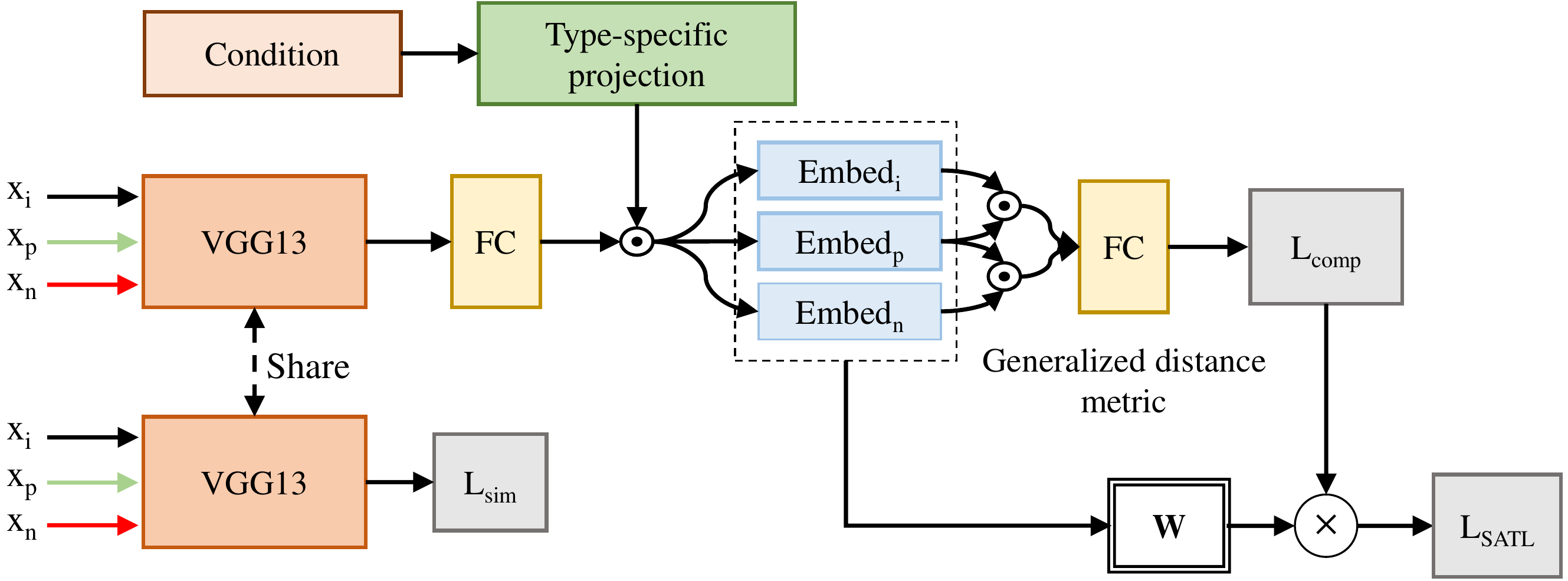}
\end{center}
   \caption{Architecture of the proposed SAT model.}
\label{fig:Architecture}
\end{figure*}

Fig.~\ref{fig:Architecture} shows the architecture of the proposed SAT model. We combine a very simple conditional similarity network with the proposed SATL to show the power of SATL in letting the CNN learn from hard items. Let us assume that we have an image $x_{i}$, a triplet input $(x_{i}^{(u)}, x_{n}^{(v)}, x_{p}^{(v)})$ is built, where $u$ and $v$ are image types. The pair $(x_{i}^{(u)}$, $x_{p}^{(v)})$ is compatible (positive) and the pair $(x_{i}^{(u)}$, $x_{n}^{(v)})$ is incompatible (negative). The triplet input is first passed into a VGG13 backbone~\cite{Simonyan_arxiv_14} to extract their general features $(F_{x_{i}}^{(u)}$, $F_{x_{n}}^{(v)}$, $F_{x_{p}}^{(v)})$. We define a type-specific embedding space $M^{(u,v)}$, in which objects of type $u$ and $v$ are matched. Associated with this space is a projection $P^{u\rightarrow(u,v)}$ which maps the embedding of an object of type $u$ to $M^{(u,v)}$. Then, for $(x_{i}^{(u)}$, $x_{p}^{(v)})$, we require the distance $\|P^{u\rightarrow(u,v)}(F_{x_{i}}^{(u)})-P^{v\rightarrow(u,v)}(F_{x_{p}}^{(v)})\|_{2}$ to be small. This does not mean $F_{x_{i}}^{(u)}$ and $F_{x_{p}}^{(v)}$ should be similar in the general space.

If the general feature dimension is $d$, our model learns a weight matrix $\textbf{w}^{(u,v)}\in{R^{d}}$. The compatibility is measured with
\begin{equation}
d_{i,j}^{(u,v)} = \|F_{x_{i}}^{(u)} \odot \textbf{w}^{(u,v)} - F_{x_{j}}^{(v)} \odot \textbf{w}^{(u,v)} \|_{2},
\label{eq:Mapping_distance}
\end{equation}
where $j$ denotes a randomly selected image from type $v$, $x_{j}$ can be compatible or incompatible with the given image $x_{i}$. $\odot$ represents component-wise multiplication.

The triplet loss of the input is calculated using:
\begin{equation}
L_{comp} =\max\{0,d_{i,p}^{(u,v)}-d_{i,n}^{(u,v)}+\mu\},
\label{eq:triplet_loss}
\end{equation}
where $\mu$ is a margin.

To let the CNN learn from hard fashion items, we develop a SATL to assign different weights to different triplet inputs,
\begin{equation}
L_{SATL} = L_{comp} \cdot \textbf{W},
\label{eq:SATL}
\end{equation}
where $\textbf{W}$ is a learned weight matrix.

We also learn the similarity loss of the general features. The loss used to learn similarity is:
\begin{equation}
L_{sim} = \max\{0,D_p-D_{n1}+\mu\}+\max\{0,D_p-D_{n2}+\mu\},
\label{eq:sim_loss}
\end{equation}
where $D_p = \|F_{x_{p}}^{(v)} - F_{x_{n}}^{(v)}\|_{2}$, $D_{n1} = \|F_{x_{p}}^{(v)} - F_{x_{i}}^{(u)} \|_{2}$, and $D_{n2} = \|F_{x_{n}}^{(v)} - F_{x_{i}}^{(u)} \|_{2}$.

Same as~\cite{Vasileva_ECCV_18}, we add an $l_{1}$ and $l_{2}$ regularization on the projection $P^{(.)\rightarrow(.,.)}$ and the learned image general features. Thus, the type-aware loss becomes,
\begin{equation}
L_{type-aware} = L_{SATL} + \lambda_{1}L_{sim} + \lambda_{2}L_{l_{1}} + \lambda_{3}L_{l_{2}},
\label{eq:type-aware_loss}
\end{equation}
where $\lambda_{1} = 5 \times 10^{-5}$, $\lambda_{2-3} = 5 \times 10^{-4}$, by following the experimental settings of~\cite{Vasileva_ECCV_18}.

\subsection{SATL}

A labeled compatible outfit can be color compatible, pattern compatible, or style compatible. It has no standard. We assume that in the type-specific embedding space $M^{(u,v)}$, the positive pair distance $d_{i,p}^{(u,v)}$ and the negative pair distance $d_{i,n}^{(u,v)}$ should meet $d_{i,n}^{(u,v)}-d_{i,p}^{(u,v)} > \mu$. If $d_{i,n}^{(u,v)}-d_{i,p}^{(u,v)} < \mu$, this triplet input is considered a hard item. To deal with the learning of hard items, we first propose a method to define the DS of each outfit,

\begin{equation}
DS_{i} = \max\{0,d_{i,p}^{(u,v)}-d_{i,n}^{(u,v)}+\mu\},
\label{eq:difficulty_score}
\end{equation}
where $\mu$ is a margin. In this paper, $\mu = 0.3$.

Then, we assign a learned weight parameter $w_{DS_{i}}$ to each triplet input. Thus, the final weight matrix becomes:

\begin{equation}
\textbf{W}_{i} = DS_{i} \cdot w_{DS_{i}}.
\label{eq:W}
\end{equation}

By assigning different weights to different triplet inputs, the DS of each triplet input could be considered, our model can better learn the compatibility among hard items, thus generating a more discriminative embedding space.

\section{Experiments}
\label{sec:expe}
With the Polyvore Outfits and Polyvore Outfits-D~\cite{Vasileva_ECCV_18} as our datasets, we conduct experiments on the popular compatibility prediction and fill-in-the-blank (FITB) tasks to evaluate our SAT model's performance. VGG13~\cite{Simonyan_arxiv_14} pre-trained on ImageNet~\cite{Deng_CVPR_09} was adopted as our backbone CNN model. The embedding size is 128, the margin $\mu$ is 0.3, and the mini-batch
size is 128. We set the initial learning rate to $5 \times 10^{-5}$. We adopt a learning rate schedule that linearly decreases the
learning rate to zero, but set the warmup ratio to zero as our initial learning rate is already small. All the losses are resigned a learned weight according to their contributions in the final prediction. Since there are no ground truth negative images for each outfit, we randomly sample a set of negative images that have the same category as the positive image similar to~\cite{Vasileva_ECCV_18}.

We compare our SAT model with state-of-the-art methods. We would like to emphasize that we executed the official codes of the previous works if available (Type-aware~\cite{Vasileva_ECCV_18}), re-implemented by ourselves if only part of the codes is available (SCE-Net~\cite{Tan_ICCV_19} and CSA-Net~\cite{Lin_CVPR_20}), and copy numbers from their papers if no codes are available (S-VAL~\cite{Kim_ICCV_21}, SSL~\cite{Revanur_RecSys_21}, and OCM-CF~\cite{Su_ACMMM_21}). Therefore, the numbers of  the Type-aware~\cite{Vasileva_ECCV_18}, SCE-Net~\cite{Tan_ICCV_19}, and CSA-Net~\cite{Lin_CVPR_20} are different from those in the original papers for a more fair comparison of later ablation studies on our SATL. The batch size is 128, the epochs are 10, the embedding size is 128, and the ResNet-18~\cite{He_CVPR_16} pre-trained on ImageNet~\cite{Deng_CVPR_09} was adopted as the CNN backbone.

Table~\ref{tab:Main_results} shows the comparison results with other state-of-the-art methods. As we can see, our SAT could achieve better performance than the state-of-the-art conditional similarity networks (Type-aware~\cite{Vasileva_ECCV_18}, SCE-Net~\cite{Tan_ICCV_19}, and CSA-Net~\cite{Lin_CVPR_20}) and some state-of-the-art methods (S-VAL~\cite{Kim_ICCV_21} and SSL~\cite{Revanur_RecSys_21}).  Even compared with the graph neural network based OCM-CF~\cite{Su_ACMMM_21}, our SAT achieves comparable performance. Besides, our SATL can be extended to other conditional similarity networks to further improve their performance.

\begin{figure*}[t]
\begin{center}
\includegraphics[width=0.7\linewidth]{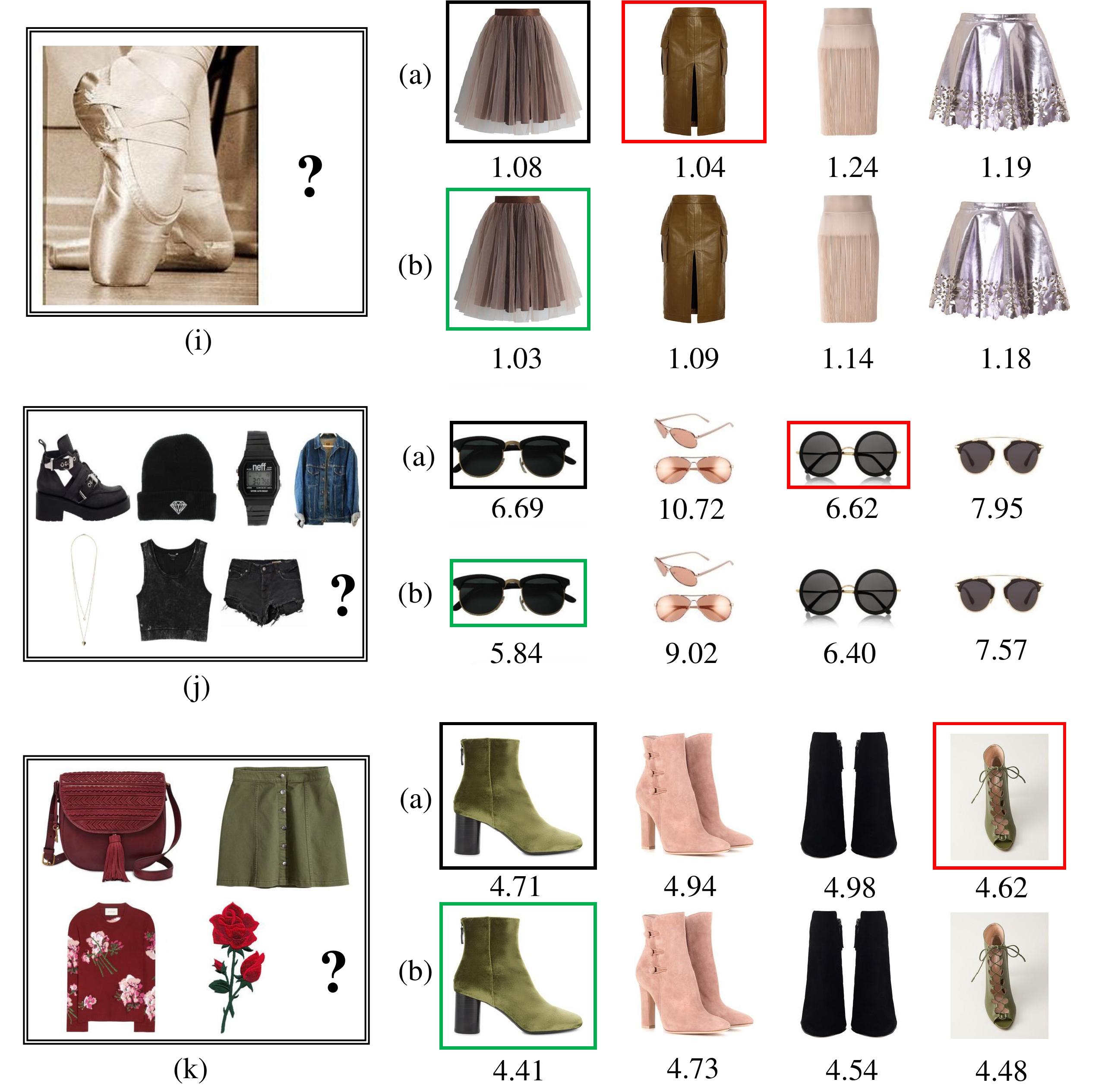}
\end{center}
   \caption{Visualization results. (i), (j), and (k) are given incomplete outfits. Row (a) are answer lists and results recommended using our model without SATL, while row (b) are answer lists and results recommended using our SAT model. The value below each image denotes its distance to the given incomplete fashion outfit, the item with a smaller value is more compatible with the given incomplete outfit compared with the item with a larger value. The item highlighted in the black box, red box, and green box is ground truth, false recommendation, and accurate recommendation, respectively.}
\label{fig:Vis}
\end{figure*}

Fig.~\ref{fig:Vis} shows some visualization examples of our SAT model. Our SAT model could accurately recommend some hard items compared with the one without SATL.

\begin{table}[t]
\caption{Comparison with other state-of-the-art methods.}
\begin{center}
\resizebox{0.48\textwidth}{!}
{
\begin{tabular}{l|ccccc}
\hline
\multirow{2}{*}{Methods}&\multirow{2}{*}{Text}& \multicolumn{2}{c}{Polyvore Outfits}& \multicolumn{2}{c}{Polyvore Outfits-D}\\
\cline{3-6}
 & & FITB Acc. &Compat. Acc. &FITB Acc. &Compat. Acc. \\\hline
Type-aware~\cite{Vasileva_ECCV_18} & Yes &57.4 &0.87 & 56.3 &0.85 \\
SCE-Net~\cite{Tan_ICCV_19} & Yes &49.9  & 0.78& 49.1 &0.76 \\
CSA-Net~\cite{Lin_CVPR_20} & No &55.8 &0.85 &  55.1&0.83\\
S-VAL~\cite{Kim_ICCV_21} & No & 55.8 &0.84 & 54.3 &0.81\\
SSL~\cite{Revanur_RecSys_21} & No & 54.9 &0.86 & 51.5 &0.82\\
OCM-CF~\cite{Su_ACMMM_21}  & No & \textbf{63.62} &\textbf{0.92} & 56.59&\textbf{0.86} \\
\textbf{Our SAT}& No& 62.2&\textbf{0.92}&\textbf{56.9} &\textbf{0.86}\\
\hline
\end{tabular}}
\end{center}
\label{tab:Main_results}
\end{table}

We also conduct ablation studies to verify the effectiveness of our SATL. Table~\ref{tab:SATL} shows that when using the SATL, both the FITB and compatibility accuracy could be improved.

\begin{table}[t]
\caption{Ablation studies of our SATL.}
\begin{center}
\resizebox{0.45\textwidth}{!}
{
\begin{tabular}{l|cccc}
\hline
\multirow{2}{*}{Methods}& \multicolumn{2}{c}{Polyvore Outfits}& \multicolumn{2}{c}{Polyvore Outfits-D}\\
\cline{2-5}
  & FITB Acc. &Compat. Acc. &FITB Acc. &Compat. Acc. \\
\hline
Without SATL& 60.9 & 0.90 & 56.0&0.85 \\
\textbf{With SATL}&\textbf{62.2}&\textbf{0.92}&\textbf{56.9} &\textbf{0.86}\\\hline
\end{tabular}}
\end{center}
\label{tab:SATL}
\end{table}

We extend our SATL to several state-of-the-art conditional similarity networks on the compatibility and FITB tasks respectively. Tables~3(a)-3(c) show the experimental results when extending our SATL to the Type-aware~\cite{Vasileva_ECCV_18}, SCE-Net~\cite{Tan_ICCV_19}, and CSA-Net~\cite{Lin_CVPR_20}. Both the compatibility and FITB accuracy of the three methods are improved when using our SATL. It demonstrates that our SATL is effective in improving the CNN's ability in learning a more discriminative embedding space in fashion compatibility prediction.

\begin{table}[t]
\centering
\caption{Extending our SATL to several state-of-the-art conditional similarity networks.}
\begin{subtable}[t]{.45\textwidth}
\centering
\caption{Extending our SATL to the Type-aware~\cite{Vasileva_ECCV_18}}
\resizebox{\textwidth}{!}
{
\begin{tabular}{l|cccc}
\hline
\multirow{2}{*}{Methods}& \multicolumn{2}{c}{Polyvore Outfits}& \multicolumn{2}{c}{Polyvore Outfits-D}\\
\cline{2-5}
& FITB Acc. &Compat. Acc. &FITB Acc. &Compat. Acc. \\
\hline
Without SATL& 57.4 & 0.87 & 56.3 & 0.85 \\
\textbf{With SATL}&\textbf{58.1}&\textbf{0.87}&\textbf{56.9 } &\textbf{0.86 }\\\hline
\end{tabular}
}
\label{tab:Type-aware}
\end{subtable}%
\hfill
\begin{subtable}[t]{.45\textwidth}
\caption{Extending our SATL to the SCE-Net~\cite{Tan_ICCV_19}}
\centering
\resizebox{\textwidth}{!}
{
\begin{tabular}{l|cccc}
\hline
\multirow{2}{*}{Methods}& \multicolumn{2}{c}{Polyvore Outfits}& \multicolumn{2}{c}{Polyvore Outfits-D}\\
\cline{2-5}
& FITB Acc. &Compat. Acc. &FITB Acc. &Compat. Acc. \\
\hline
Without SATL&49.9  & 0.78& 49.1 &0.76 \\
\textbf{With SATL}&\textbf{50.8}&\textbf{0.78}&\textbf{50.1} &\textbf{0.77}\\\hline
\end{tabular}
}
\label{tab:SCE-Net}
\end{subtable}%
\hfill
\begin{subtable}[t]{.45\textwidth}
\caption{Extending our SATL to the CSA-Net~\cite{Lin_CVPR_20}}
\centering
\resizebox{\textwidth}{!}
{
\begin{tabular}{l|cccc}
\hline
\multirow{2}{*}{Methods}& \multicolumn{2}{c}{Polyvore Outfits}& \multicolumn{2}{c}{Polyvore Outfits-D}\\
\cline{2-5}
& FITB Acc. &Compat. Acc. &FITB Acc. &Compat. Acc. \\
\hline
Without SATL&55.8 &0.85 & 55.1&0.83 \\
\textbf{With SATL}&\textbf{56.7}&\textbf{0.86}&\textbf{56.0} &\textbf{0.84}\\\hline
\end{tabular}
}
\label{tab:CSA-Net}
\end{subtable}%
\label{tab:ablation_studies}
\end{table}

\section{Conclusions}
\label{sec:conclu}

In this paper, we proposed an SAT model for fashion compatibility prediction, which can effectively enhance the discriminative power of embedding space for hard item recommendations. To the best of our knowledge, this is the first work to design a method letting the CNN learn more from the hard items in fashion compatibility prediction.  We first designed a method to define hard outfits and a
DS was defined and assigned to a triplet input which was correspondent to the difficulty in recommending an item for this outfit. Then, we developed a SATL that combines the standard triplet loss with an adaptive weight generated based on the DS of fashion items. Finally, combining SATL, we proposed a very simple SAT model which only uses image features, but could perform better than the state-of-the-art methods.
We extended our SATL to other conditional similarity networks and the experimental results demonstrated that our SATL is very effective in improving other models' performance.


\end{document}